\documentclass[10pt]{article}
\usepackage[preprint]{tmlr}

\usepackage{hyperref}
\usepackage{url}
\usepackage{amsmath,amssymb}
\usepackage{graphicx}
\usepackage{booktabs}
\usepackage{multirow}

\title{Weber's Law in Transformer Magnitude Representations:\\
Efficient Coding, Representational Geometry, and Psychophysical Laws in Language Models}

\author{\name Jon-Paul Cacioli \\
      \addr Independent Researcher, Melbourne Australia}

\def\month{MM}
\def\year{YYYY}
\def\openreview{\url{https://openreview.net/forum?id=XXXX}}

\begin{document}

\maketitle

\begin{abstract}
How do transformer language models represent magnitude? Recent work disagrees: some find logarithmic spacing, others linear encoding, others per-digit circular representations. We argue that the disagreement is methodological and apply the formal tools of psychophysics to resolve it. Using four converging paradigms (representational similarity analysis, behavioural discrimination, precision gradients, causal intervention) across three magnitude domains (numerical, temporal, spatial) in three 7--9B instruction-tuned models spanning three architecture families (Llama, Mistral, Qwen), we report three findings. \textbf{First,} representational distance structure is consistently log-compressive: RSA correlations with a Weber-law dissimilarity matrix ranged from .68 to .96 across all 96 model--domain--layer cells (2 models $\times$ 3 domains $\times$ 16 primary layers), with both cosine and Euclidean metrics agreeing and linear geometry never preferred. \textbf{Second,} this geometry is dissociated from behaviour: one model produces a human-range Weber fraction (WF = 0.20, i.e. $\sim$20\% relative difference threshold) while the other does not, and both models perform at chance on temporal and spatial discrimination despite possessing logarithmic geometry for those domains. \textbf{Third,} causal intervention reveals a layer dissociation: early layers where geometry is weakest are functionally implicated in approximate magnitude processing (4.1$\times$ specificity), while later layers where geometry is strongest are not causally engaged under our linear intervention (1.2$\times$). Corpus analysis confirms the efficient coding precondition: integer frequencies in the training data follow a power law ($\alpha$ = 0.77). These results are consistent with the hypothesis that training data statistics (without biological resource constraints) are sufficient to produce log-compressive magnitude geometry, but that geometry alone does not guarantee behavioural competence.
\end{abstract}

\section{Introduction}

Transformer language models process numerical, temporal, and spatial magnitudes as part of ordinary language understanding. Recent work has begun to characterise how these magnitudes are internally represented, but the emerging picture is contradictory. \citet{alquabeh2025number} extracted principal components from hidden states and found logarithmic spacing. \citet{zhu2025language} trained linear probes and concluded the encoding is linear. \citet{levy2025language} demonstrated per-digit circular representations in base 10, suggesting that apparent magnitude geometry is an artefact of digit-level structure. \citet{kadlcik2025unravelling} found convergent sinusoidal representations across model families, pointing to a Fourier decomposition. \citet{veselovsky2025what} showed that number representations entangle Levenshtein string edit distance with numerical distance.

These findings are not necessarily contradictory. They may reflect different methodological windows onto the same underlying representation. But the literature lacks a principled framework for characterising the functional form of magnitude representations and their relationship to behaviour. We propose that psychophysics provides this framework.

Psychophysics is the quantitative study of the relationship between stimulus magnitudes and their mental representations. Its central law, Weber's Law (\citealt{weber1834pulsu}), states that the just-noticeable difference between two stimuli is proportional to their magnitude: $\Delta$$S$/$S$ = $k$. \citet{moyer1967time} first demonstrated that this ratio-dependent regularity extends to symbolic number comparison in humans: reaction times for judging which of two digits is larger decrease with their numerical distance, consistent with an analogue magnitude representation obeying Weber's Law. In modern cognitive neuroscience, this regularity is understood as a consequence of \emph{efficient coding}: when a system with limited capacity encodes a variable whose natural distribution follows an approximate 1/$s$ power law, the information-maximising encoding is logarithmic, and logarithmic encoding produces Weber-like discrimination (\citealt{ganguli2014efficient}; \citealt{laughlin1981simple}). The efficient coding framework makes a precise, testable prediction: if the input distribution is scale-invariant and the system optimises an information-theoretic objective, the encoding should be logarithmic.

\textbf{Key psychophysical concepts for ML readers.} A \emph{Weber fraction} (WF) is the ratio $\Delta$S/S at which a subject can discriminate two magnitudes at threshold (e.g., WF = 0.20 means a $\sim$20\% relative difference is needed). \emph{Stevens' exponent} $\beta$ characterises the power-law compression of an internal representation: $\beta$ = 1 is linear, $\beta$ $\to$ 0 is logarithmic, and 0 < $\beta$ < 1 is intermediate compression. \emph{Representational similarity analysis} (RSA; \citealt{kriegeskorte2008representational}) compares a model's pairwise distance matrix (the representational dissimilarity matrix, or RDM) to theoretical prediction matrices using rank correlation. A \emph{Mantel test} assesses RSA significance via permutation, accounting for the non-independence of pairwise distances that share endpoints, which is critical when the RDM has C($n$, 2) entries but only $n$ independent observations.

Transformers meet the distributional precondition. Numerical magnitudes in natural text follow approximately Benford's Law, with small numbers vastly more frequent than large ones (\citealt{dehaene1992cross}). But transformers lack the biological constraint: there is no metabolic cost, no firing-rate bound, no fixed population size. This creates a natural experiment. If data statistics alone are sufficient to produce logarithmic representations under gradient-based optimisation, the models should develop Weber-like magnitude geometry. If biological constraints are necessary, they should not. \textbf{We frame this as a test of functional convergence:} do different optimisation regimes, operating on input with the same distributional structure, converge on the same representational geometry?

Critically, we do not merely ask whether representations are ``logarithmic or linear'', a question that existing methods can answer. We ask a richer set of questions that psychophysics was designed for: \emph{Does the geometry translate into ratio-dependent behaviour? Does the system allocate more representational precision to small magnitudes? Is the magnitude subspace causally relevant to comparison? Does the encoding generalise across magnitude domains?} These questions require the methodological toolkit of psychophysics: Weber fractions, psychometric functions, representational similarity analysis, precision gradients, and causal intervention.

\subsection{Contributions}

Our contributions are as follows:

\textbf{1.} A psychophysics-grounded framework for probing magnitude representations. We apply Weber fractions, psychometric functions, RSA with Mantel permutation tests, precision gradients, and causal intervention to LLM hidden states. To our knowledge, this is the first application of the full psychophysical toolkit to this problem, with pre-registered hypotheses and all results reported regardless of outcome.

\textbf{2.} Empirical resolution of conflicting findings. Logarithmic geometry is consistent across all 96 model--domain--layer cells tested (both distance metrics agreeing), with linear geometry never preferred, resolving the apparent \citeauthor{alquabeh2025number}--\citeauthor{zhu2025language} contradiction. Corpus analysis confirms the efficient coding precondition ($\alpha$ = 0.77). Variance partitioning quantifies the digit-encoding confound (\citealt{levy2025language}) as real but subordinate to magnitude compression.

\textbf{3.} Three informative dissociations. (a) Geometry without behavioural competence: both models possess logarithmic temporal and spatial geometry but perform at chance on discrimination tasks for those domains. (b) Model-specific readout: Llama produces a human-range Weber fraction; Mistral does not. (c) Causal layer inversion: early layers are functionally implicated in approximate comparison; later layers where geometry is strongest are not causally engaged under our patching intervention.

\section{Related Work}

\subsection{Number Representations in Language Models}

\textbf{Representation geometry.} \citet{alquabeh2025number} used PCA and partial least squares to show that LLM hidden states embed numbers on a logarithmically compressed manifold. \citet{zhu2025language} reached the opposite conclusion using linear probes, finding that probe accuracy was best predicted by linear, not logarithmic, target functions. We show that both findings are compatible: a smooth logarithmic encoding is globally logarithmic (\citeauthor{alquabeh2025number} is right about the geometry) while being locally linear enough for linear probes to succeed (\citeauthor{zhu2025language} is right that probes recover values). The key distinction is methodological: global geometry measures (RSA, model comparison) versus local recovery measures (probe accuracy). For a comprehensive taxonomy of number encoding methods in NLP, see \citet{thawani2021representing}. \citet{razeghi2022impact} demonstrated that pretraining term frequencies strongly predict LLM numerical reasoning accuracy, motivating our corpus distribution analysis and frequency-matched controls.

\textbf{Digit-level structure.} \citet{levy2025language} demonstrated that LLMs encode multi-digit numbers as per-digit circular representations in base 10, not as unified magnitudes. This is the most significant threat to interpreting apparent magnitude geometry at face value. \citet{veselovsky2025what} showed further that LLM number representations entangle string edit distance with numerical distance. Our digit-boundary diagnostic and variance partitioning address this directly: we find that digit structure is a real feature of the representation (Cohen's $d$ > 0.8 at most layers) but is subordinate to magnitude compression (log-magnitude explains 55--63\% of RDM variance versus 8--15\% for digit-count).

\textbf{Fourier and trigonometric encodings.} \citet{kadlcik2025unravelling} found that different LLM families converge on similar sinusoidal number representations, suggesting a Fourier decomposition. We pre-registered a residual periodicity analysis (E4) to test this alternative, triggered if our three-model framework (linear, logarithmic, power-law) fitted poorly ($R^2$ < .20). The trigger was never activated: all cells exceeded .68, indicating that the logarithmic/power-law framework captures the dominant structure.

\subsection{Efficient Coding and Weber's Law}

The efficient coding framework (\citealt{barlow1961possible}; \citealt{laughlin1981simple}; \citealt{ganguli2014efficient}) explains Weber's Law as a consequence of information-maximising encoding under natural stimulus statistics. When the prior distribution is approximately 1/$s$ (scale-invariant), the optimal encoding is logarithmic, producing constant discrimination thresholds in ratio units, producing Weber's Law. This has been verified in biological systems from fly photoreceptors (\citealt{laughlin1981simple}) to primate numerosity tuning (\citealt{nieder2003coding}). \citet{pardo2019mechanistic} demonstrated the mechanistic foundation in rodent somatosensory cortex. We test whether the same representational signature emerges in a system that shares the distributional precondition (1/$s$ input) but not the biological constraint (metabolic cost).

\subsection{Psychophysics Applied to Language Models}

\citet{marjieh2023large} showed that LLMs predict human sensory judgements across six modalities, treating LLMs as models of human perception. \citet{gurnee2024language} demonstrated that LLMs represent space and time in their activations but did not test whether these representations follow psychophysical laws. We ask the complementary question: not whether LLMs predict human psychophysics, but whether they \emph{exhibit} it: whether the internal representations themselves have the structural properties (logarithmic compression, ratio-dependent discrimination, graded precision) that characterise magnitude representations in the cognitive science literature.

\section{Method}

All hypotheses, success criteria, analysis plans, stimulus files, and decision rules were pre-registered on the Open Science Framework prior to data collection (https://osf.io/u4wp5). The primary registration document (v2.7) and a supplementary amendment (v2.8, documenting three infrastructure changes) are publicly available. All analysis code is archived at https://anonymous.4open.science/r/weber-B02C.

\subsection{Models}

Two instruction-tuned models served as primary participants: \textbf{Llama-3-8B-Instruct} (8.03B parameters, 32 transformer layers, $d_{\text{model}}$ = 4096) and \textbf{Mistral-7B-Instruct-v0.3} (7.24B parameters, 32 layers, $d_{\text{model}}$ = 4096). Both were loaded in FP16 precision via HuggingFace Transformers (v5.3) with PyTorch ROCm 6.4 on an AMD Radeon RX 7900 GRE GPU (16 GB VRAM). \textbf{Llama-3-8B} (base, non-instruct) served as an exploratory comparison for investigating instruction tuning. \textbf{Qwen-2.5-7B-Instruct} (Alibaba; 7.6B parameters, 28 layers, $d_{\text{model}}$ = 3584) was added post-registration as a second exploratory model to test cross-architecture generalisability; it represents a third architecture family with distinct training data. Cross-precision verification confirmed FP16--BF16 agreement (worst Pearson $r$ = .9999 across 16 primary layers). Exact commit hashes are recorded for reproducibility.

\subsection{Stimuli and Magnitude Domains}

Three magnitude domains were tested, with numerical as the primary domain.

\textbf{Numerical.} 26 probing values spanning three orders of magnitude: \{1, 2, ..., 9, 10, 15, 20, 30, ..., 90, 100, 150, 200, 300, 500, 700, 1000\}. Presented as Arabic digits in five carrier sentences (e.g., ``The number {[}N{]} is a quantity''). 25/26 values tokenise as single tokens in Llama-3; all 26 are multi-token in Mistral (leading space + per-digit tokens). 1,500 comparison pairs: 5 baselines $\times$ 6 ratios (1.05--3.00) $\times$ 50 pairs, with jittered baselines ($\pm$15\%, seed 42).

\textbf{Temporal.} 19 probing values spanning $\sim$7 orders of magnitude in seconds (1 second to 1 year), in natural mixed units. 900 comparison pairs.

\textbf{Spatial.} 14 probing values spanning $\sim$6 orders of magnitude (1 metre to 1000 km), in natural mixed units. 900 comparison pairs.

\subsection{Four Converging Paradigms}

\begin{table}[t]
\caption{Overview of the four experimental paradigms.}
\label{tab:paradigms}
\centering
\footnotesize
\begin{tabular}{p{2.5cm}p{3.5cm}p{3.5cm}p{3.5cm}}
\toprule
\textbf{Paradigm} & \textbf{Measures} & \textbf{Dependent variable} & \textbf{Falsification} \\
\midrule
A: Representational Geometry Probing & Hidden-state distances between magnitude encodings & Cosine/Euclidean distance in activation space & Linear geometry refutes Weber \\
B: Behavioural Discrimination & Forced-choice magnitude comparison accuracy & Proportion correct, Weber fraction & Accuracy predicted by absolute diff, not ratio \\
C: Precision Gradient & How representational density changes with magnitude & Local pairwise distance between adjacent magnitudes & Flat precision gradient refutes efficient coding \\
D: Causal Intervention & Whether magnitude subspace is functionally relevant & $\Delta$p after activation patching & No causal effect means geometry is epiphenomenal \\
\bottomrule
\end{tabular}
\end{table}

\subsubsection{Paradigm A: Representational Geometry Probing}

For each probing sentence, the model's forward pass was run with hidden-state extraction at all 33 layers (embedding + 32 transformer layers). Hidden-state vectors were extracted at the magnitude token position (single-token) or final token of the magnitude expression (multi-token), verified against character-to-token offset mappings for all 295 probing sentences per model (zero mismatches).

Vectors were averaged across 5 carrier sentences to obtain centroid representations per magnitude per layer. Pairwise distances were computed for all $C$(26, 2) = 325 pairs using both \textbf{cosine distance} and \textbf{Euclidean distance} as co-primary metrics. Three geometric models were fitted: \textbf{Linear} ($d$ = $a$ + $b$$\cdot$$|$$n_1$ -- $n_2$$|$), \textbf{Weber/Log} ($d$ = $a$ + $b$$\cdot$$|$ log($n_1$) -- log($n_2$)$|$), and \textbf{Stevens/Power} ($d$ = $a$ + $b$$\cdot$$|$$n_1$$^\beta$ -- $n_2$$^\beta$$|$, $\beta$ estimated by NLS). Models compared via $R^2$ and AIC.

\textbf{Representational Similarity Analysis (RSA).} At each layer, the model's representational dissimilarity matrix (RDM) was compared to three theoretical RDMs (Linear, Weber, Stevens) via Spearman rank correlation. Significance was assessed with Mantel permutation tests (10,000 permutations), which account for the non-independence of pairwise distances sharing endpoints (\citealt{kriegeskorte2008representational}). Theoretical RDMs were z-scored before computing correlations.

\subsubsection{Paradigm B: Behavioural Magnitude Discrimination}

Three approximate comparison tasks that require magnitude processing rather than digit-parsing. \textbf{B1 (cross-format, primary):} ``Which represents a larger quantity: A) {[}expression$_1${]} or B) {[}expression$_2${]}?'' (e.g., ``three dozen'' vs. ``twenty-eight''). \textbf{B2 (approximate arithmetic):} ``Without calculating exactly, which is larger?'' (e.g., ``47\% of 89'' vs. ``38\% of 112''). \textbf{B3 (contextual):} a brief context mentioning two quantities. A \textbf{symbolic control} presented Arabic digit pairs, predicted at ceiling. All scoring via greedy decoding (T = 0) with logit extraction. Entropy diagnostic flags exact vs. approximate processing.

\subsubsection{Paradigm C: Representational Precision Gradient}

Local precision defined as 1/$|$$|$$h$($n$+1) -- $h$($n$)$|$$|$. Efficient coding predicts this decreases with magnitude. Under logarithmic geometry with non-uniform stimulus spacing, precision normalised by log step size should be constant; we report both raw and normalised precision.

\subsubsection{Paradigm D: Causal Intervention}

A ridge-regression probe trained on Paradigm A hidden states predicting log(magnitude) defined the magnitude direction $v_{\text{mag}}$. For 200 comparison prompts, additive activation patching along $v_{\text{mag}}$ was applied at four dose levels (0.25--1.00), preserving all orthogonal components. Change in comparison probability ($\Delta$$p$) was recorded. 10 random-direction controls established the null distribution.

\textbf{Direction validation.} Because the probe's dimensionality (4096) far exceeds the number of magnitude values (26), probe $R^2$ = 1.0 at all layers, making layer selection via probe fit degenerate. We validated the magnitude direction in two ways. First, PCA was applied to the 26 magnitude centroids at each layer; PC1 was required to correlate with log(magnitude) at $r$ > .80 to confirm that the dominant axis of variation aligns with the probe direction. Second, the E5 exploratory analysis tested causal effects at multiple layers (not just the probe-selected layer), providing a layer-level causal profile that does not depend on probe-based layer selection. The specificity control (magnitude direction vs. 10 random directions) further ensures that any observed effects are attributable to the magnitude subspace specifically, not to arbitrary high-dimensional directions.

\subsection{Robustness Controls}

Six pre-registered controls: (1) \textbf{digit-boundary diagnostic} + variance partitioning of log-magnitude vs. digit-count contributions to the RDM; (2) \textbf{single-token control} verifying geometry is not a tokenisation artefact; (3) \textbf{unit-boundary check} testing cross-unit magnitude abstraction; (4) \textbf{cross-precision verification} (FP16 vs. BF16); (5) \textbf{frequency-matched noun control} (26 nouns matched by log-probability via Hungarian algorithm, Spearman $\rho$ = .89) with a \textbf{semantic structure diagnostic} disambiguating frequency from meaning; (6) \textbf{shuffled-magnitude sanity check} testing whether geometry tracks token identity or carrier context.

\subsection{Corpus Magnitude Distribution}

All integer mentions (1--1000) were extracted from 500,000 OpenWebText documents. Power-law ($f$($n$) $\propto$ $n$$^{-\alpha}$) and exponential models were fitted via AIC, with Benford's Law compliance analysis.

\subsection{Hypotheses and Statistical Analysis}

Three primary hypotheses at $\alpha$ = .017 (Bonferroni for 3): \textbf{H1} (log geometry via RSA Mantel + AIC at $\geq$9/17 primary layers in $\geq$2/3 domains); \textbf{H2} (behavioural Weber's Law via $\Delta$ deviance test on B1); \textbf{H3} (negative precision gradient at $\geq$17/32 layers in $\geq$2 domains). All require both models for programme-level support. Four secondary hypotheses (H4--H7) at $\alpha$ = .05, evaluated per-model. Pre-registered exploratory analyses: E1 (base vs. instruct), E2 (Stevens power law), E3 (bridge to SDT), E4 (residual periodicity, conditional), E5 (layer-specific causal profiles).

\section{Results}

Three principal findings emerged from the pre-registered analyses. First, representational geometry was consistently logarithmic across all tested conditions, resolving the log-vs-linear debate. Second, this geometry was dissociated from behavioural competence. It was present for all three magnitude domains, but only translated into Weber-like discrimination for numerical magnitudes and only in one of two models. Third, causal intervention revealed a layer inversion: early layers were functionally implicated in approximate comparison, while later layers with the strongest geometry were not causally engaged. We report each finding in detail, followed by robustness controls.

\subsection{Corpus Distribution Confirms the Efficient Coding Precondition}

Analysis of 6,046,515 integer mentions from OpenWebText yielded a power-law exponent $\alpha$ = 0.773 (the 1/$s$ prior corresponds to $\alpha$ $\approx$ 1.0). Leading-digit frequencies complied with Benford's Law within 1--2 percentage points. The distributional precondition for the efficient coding prediction (approximate scale invariance) holds in the training data. The modest deviation from the ideal $\alpha$ = 1 does not undermine the prediction. The \citet{ganguli2014efficient} framework produces approximately logarithmic compression for any power-law prior with $\alpha$ in the range 0.5--1.5.

\subsection{H1: Representational Geometry Is Consistently Logarithmic (Supported)}

\textbf{H1 was supported in both models across all tested conditions.} Both models, all three domains, both distance metrics, and all 16 primary layers (transformer layers 16--31; the pre-registration specified ``layers 16--32'' but the 32-layer models have transformer layers indexed 0--31, yielding 16 layers in the primary range; the pre-registered threshold of $\geq$9 layers is exceeded in all cases) favoured logarithmic over linear geometry. Table~\ref{tab:h1} summarises the layer-level pass rates.

\begin{table}[t]
\caption{H1 (logarithmic geometry) results: layers passing / 16 primary layers. Pass criterion: Weber RSA $\rho$ > Linear RSA $\rho$ (Mantel $p$ < .017) AND Weber AIC < Linear AIC.}
\label{tab:h1}
\centering
\footnotesize
\begin{tabular}{llll}
\toprule
& \textbf{Numerical} & \textbf{Temporal} & \textbf{Spatial} \\
\midrule
Llama (cosine) & 16/16 & 16/16 & 16/16 \\
Llama (Euclidean) & 16/16 & 16/16 & 16/16 \\
Mistral (cosine) & 16/16 & 16/16 & 16/16 \\
Mistral (Euclidean) & 16/16 & 16/16 & 16/16 \\
\bottomrule
\end{tabular}
\end{table}

At peak layers (Llama numerical, cosine), Weber $R^2$ = .83--.88, compared with Linear $R^2$ = .14--.32. RSA Spearman $\rho$ for the Weber RDM was .93--.96, versus .51--.69 for Linear, a consistent advantage of $\sim$.35 correlation units. Mantel permutation tests were significant ($p$ < .001) at all primary layers for all model--domain combinations. The cosine and Euclidean co-primary metrics were in complete agreement: both produced identical 16/16 pass rates across all model--domain cells, and no instance of metric disagreement on the best-fitting model was observed. \textbf{Linear geometry was never the best-fitting model at any layer for any domain in either model.}

AIC model selection confirmed the RSA result. At primary layers, the Weber model achieved a lower AIC than the Linear model at every layer for numerical and spatial domains in both models. For temporal duration, Stevens competed (Llama: Weber 7, Stevens 9 of 16 layers; Mistral: Weber 12, Stevens 4), consistent with compression that does not fully reach logarithmic for time representations. Linear achieved the lowest AIC at zero layers across all cells. Carrier sentence consistency was high: mean ICC across primary layers was .95--.99 for all model--domain combinations (Llama temporal .97, spatial .99; Mistral numerical .95, temporal .96, spatial .99), confirming that the geometry is robust across carrier contexts.

The Stevens exponent $\beta$ provides a continuous compression measure. At primary layers, $\beta$ converged to approximately 0.01 for numerical representations in both models, representing effectively pure logarithmic compression. This exceeds the compression observed in any human sensory modality (human $\beta$ ranges from $\sim$0.33 for brightness to $\sim$3.5 for electric shock; \citealt{stevens1961honor}), consistent with the absence of metabolic constraints that limit biological encoding resolution. An architectural explanation may also contribute: the model must represent three orders of magnitude (1--1000) within a shared 4096-dimensional embedding space alongside all other semantic features, creating a representational bottleneck that favours aggressive compression.

\textbf{Model-fit floor check and E4 (Fourier alternative).} The pre-registered residual periodicity analysis (E4), designed to test \citeauthor{kadlcik2025unravelling}'s (\citeyear{kadlcik2025unravelling}) Fourier encoding hypothesis, was contingent on any model--domain cell showing $R^2$ < .20 at a majority of primary layers. This trigger was never activated: the minimum $R^2$ across all 96 cells was .68. The logarithmic/power-law framework captured the dominant structure of magnitude geometry, and there was no evidence of residual periodic encoding that the three-model framework failed to explain.

\begin{figure}[t]
\centering
\includegraphics[width=\textwidth]{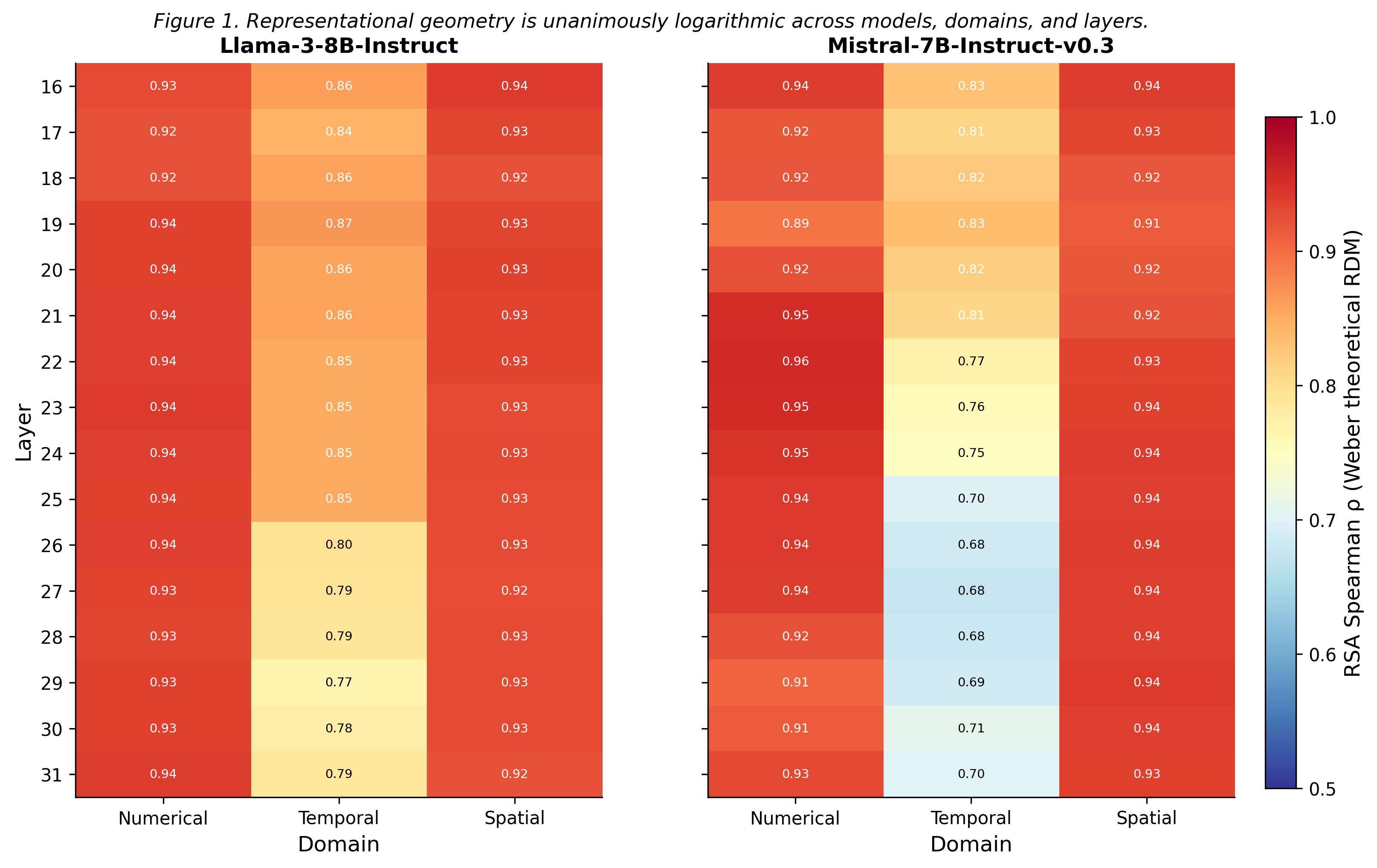}
\caption{Representational geometry is unanimously logarithmic across models, domains, and layers.}

\label{fig:figure1}
\end{figure}

\subsection{H2: Behavioural Weber's Law Is Model-Specific (Not Supported at Programme Level)}

\emph{Table~\ref{tab:behavioural}. Paradigm B results across domains and task types.}

\begin{table}[t]
\caption{Paradigm B results across domains and task types.}
\label{tab:behavioural}
\centering
\footnotesize
\begin{tabular}{p{2.2cm}p{1.8cm}p{1.8cm}p{2.8cm}p{2.8cm}}
\toprule
\textbf{Domain / Task} & \textbf{Llama accuracy} & \textbf{Mistral accuracy} & \textbf{Llama ratio effect} & \textbf{Mistral ratio effect} \\
\midrule
Numerical B1 (primary) & 78.5\% & 74.8\% & $\checkmark$ Weber-like & Weak (abs diff wins) \\
Numerical B2 & 57.9\% & 69.0\% & $\times$ Chance & $\times$ Chance \\
Numerical B3 & 90.6\% & 86.3\% & $\checkmark$ Strong & $\checkmark$ Yes \\
Symbolic control & 99.9\% & 50.0\% & $\times$ Ceiling & $\times$ Position bias \\
Temporal B1 & 47.9\% & 47.1\% & $\times$ Chance & $\times$ Chance \\
Spatial B1 & 49.7\% & 50.0\% & $\times$ Chance & $\times$ Chance \\
Base model B1 & 50.0\% & --- & $\times$ Can't instruct-follow & --- \\
\bottomrule
\end{tabular}
\end{table}

On Task B1, Llama showed monotonic accuracy increase from 75.2\% (ratio 1.05) to 89.2\% (ratio 3.00): $\Delta$ deviance = 7.51, $p$ = .006. Position-corrected Weber fraction: WF = 0.201 (BCa 95\% CI {[}0.039, 0.360{]}). The point estimate falls within the human literature range for numerical magnitude discrimination (0.10--0.25; \citealt{dehaene2003neural}), though the confidence interval is wide and extends beyond this range on both sides, reflecting the limited number of baseline levels. The pre-registered constancy criterion was met via the second arm: the aggregate WF falls within the human range (Cochran's Q was not computed because this condition was satisfied directly). Mistral's behaviour was better explained by absolute difference ($\Delta$ deviance = --12.07), yielding WF = 0.529, outside the human range.

\textbf{Temporal and spatial: geometry without competence.} Both models performed at chance on temporal and spatial B1 despite possessing logarithmic geometry for these domains (RSA $\rho$ as strong as for numerical). This is the study's most informative dissociation: representational geometry is necessary but not sufficient for behavioural magnitude discrimination. The unit-boundary control (Section~\ref{sec:controls}) explains why.

Entropy diagnostics revealed a processing-mode divergence: Llama's B1 mean logit entropy was 0.288 (above the 0.20 threshold indicating approximate processing), while Mistral's was 0.225 (flagged as closer to exact processing). This may explain why Mistral's comparison behaviour tracks absolute difference rather than ratio: it may attempt exact conversion rather than approximate magnitude comparison.

\textbf{Mistral's symbolic control failure (49 of 1500 B1 trials, 3.3\%, were excluded because Mistral's greedy output was not a valid A or B token; Llama produced valid responses on all 1500 trials).} Mistral's 50.0\% accuracy on the symbolic comparison control (pure position bias) is itself a noteworthy finding. Even on exact Arabic digit comparisons (e.g., ``Which is larger, 47 or 83?''), Mistral could not reliably access numerical magnitude for comparison. This is not a formatting issue. The same chat template and A/B labelling was used for both models. The deficit suggests that Mistral-7B-Instruct-v0.3's instruction tuning did not produce a functional comparison readout for numerical magnitude, despite the presence of the same logarithmic geometry as Llama.

\textbf{B2 task (approximate arithmetic).} B2 accuracy was 57.9\% (Llama) and 69.0\% (Mistral), above chance but showing a reversal of the B1 pattern: Mistral outperformed Llama. This may reflect Mistral's lower entropy (more exact) processing style: if B2's percentage expressions are amenable to exact computation (e.g., ``47\% of 89'' $\approx$ 41.83), Mistral's exact-processing approach may yield better results than Llama's approximate processing. Neither model showed a meaningful ratio effect on B2, confirming that approximate arithmetic remains beyond these models' approximate magnitude capabilities.

\textbf{H4 (cross-domain Weber fraction).} H4 tested whether Weber fractions differ across domains. Because both models performed at chance on temporal and spatial B1, domain-specific Weber fractions could not be estimated for non-numerical domains. H4 is therefore non-evaluable as pre-registered.

\begin{figure}[t]
\centering
\includegraphics[width=\textwidth]{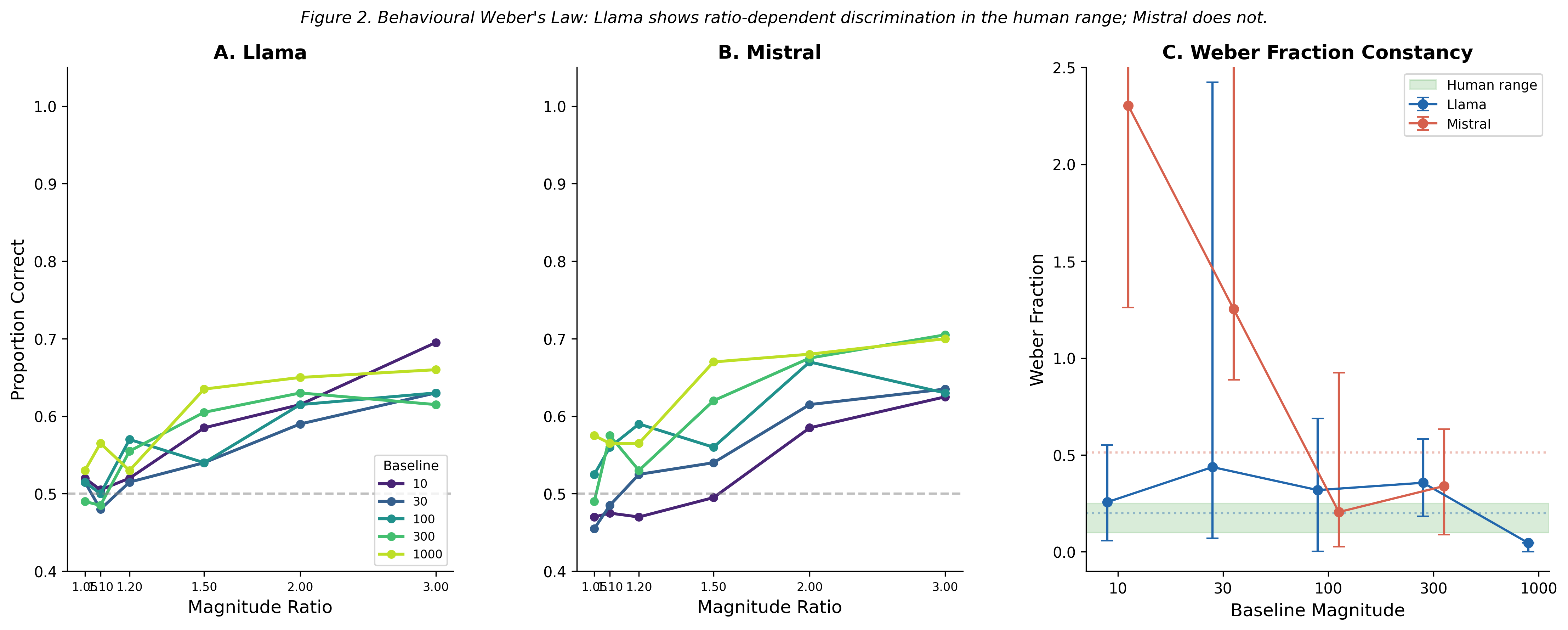}
\caption{Behavioural Weber's Law: Llama shows ratio-dependent discrimination in the human range; Mistral does not.}

\label{fig:figure2}
\end{figure}

\subsection{H3: Precision Gradient (Not Supported at Programme Level)}

Raw precision showed the predicted negative gradient for Llama numerical (31/33 layers significant) and both models' temporal domains, but failed for spatial and for Mistral numerical. H3 is therefore not supported at the programme level, and we report this as a null result on the pre-registered criterion. As an exploratory observation, log-normalised precision was approximately flat across all models and domains (power-law exponent $\gamma$ $\approx$ 0.002--0.048). This flatness is \emph{consistent with} logarithmic geometry given non-uniform stimulus spacing. Log geometry predicts that raw precision decreases with magnitude, and normalisation by log step size should remove this decrease. However, the normalised-precision interpretation was not part of the pre-registered success criterion for H3, and we flag it as exploratory to avoid the appearance of moving the goalposts. The failure of the raw gradient for spatial representations and for Mistral numerical remains unexplained and may reflect domain-specific or model-specific limitations of the efficient coding account.

\begin{figure}[t]
\centering
\includegraphics[width=\textwidth]{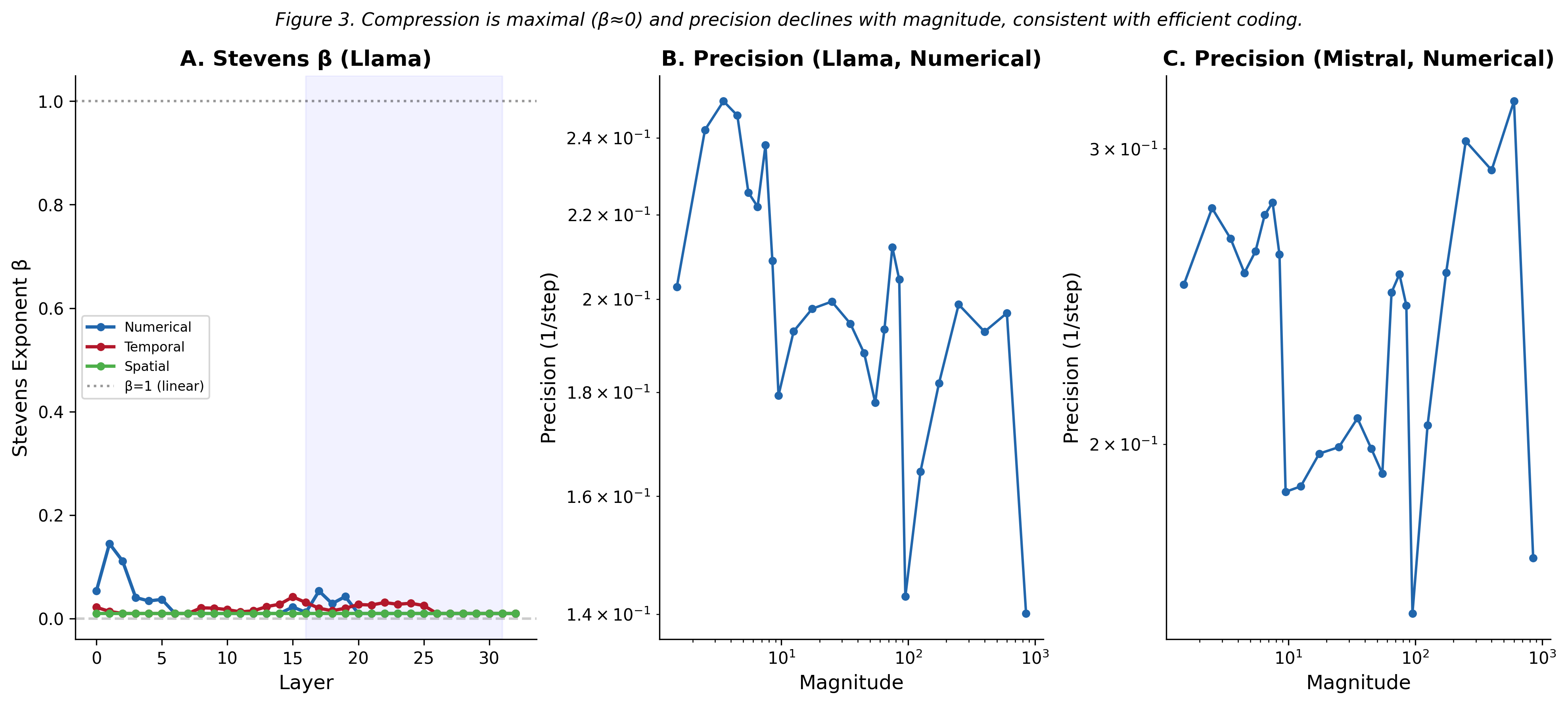}
\caption{Compression is maximal ($\beta \approx 0$) and precision declines with magnitude, consistent with efficient coding.}

\label{fig:figure3}
\end{figure}

\subsection{Secondary Hypotheses}

\emph{Table~\ref{tab:hypotheses}. Summary of all hypothesis tests.}

\begin{table}[t]
\caption{Summary of all hypothesis tests.}
\label{tab:hypotheses}
\centering
\footnotesize
\begin{tabular}{p{2.5cm}p{4cm}p{4cm}p{2.5cm}}
\toprule
\textbf{Hypothesis} & \textbf{Llama} & \textbf{Mistral} & \textbf{Programme level} \\
\midrule
H1 (log geometry) & PASS (16/16 layers, 3/3 domains) & PASS (16/16 layers, 3/3 domains) & SUPPORTED \\
H2 (behavioural Weber) & PASS ($\Delta$dev=7.51, WF=0.201) & FAIL ($\Delta$dev=--12.07, WF=0.529) & NOT SUPPORTED \\
H3 (precision gradient) & PASS (num, temp) & Marginal (temp only) & NOT SUPPORTED \\
H4 (cross-domain WF) & Non-evaluable (temp/spat at chance) & Non-evaluable & NON-EVALUABLE \\
H5 (layer transition) & PASS (num $\rho$=--.47, spat $\rho$=--.35) & PASS (num only) & NOT SUPPORTED \\
H6 (dist + ratio effects) & Partial (dist $\checkmark$, ratio $\times$, inter. $\checkmark$) & Partial (dist $\checkmark$, ratio $\checkmark$, inter. $\times$) & NOT SUPPORTED \\
H7 (causal, symbolic) & FAIL (51.5\%, need 75\%) & FAIL (43.0\%) & NOT SUPPORTED \\
\bottomrule
\end{tabular}
\end{table}

\textbf{H7} failed because the pre-registered paradigm used symbolic comparison prompts, where both models achieved near-ceiling accuracy (97.6\% and 99.4\%), leaving no room for patching to shift responses. This ceiling effect motivated exploratory analysis E5.

\subsection{Exploratory Findings}

\textbf{E1: Geometry is a pretraining property.} The base model's numerical geometry was at least as strong as the instruct model's (Weber $R^2$ = .88 vs. .86; RSA $\rho$ = .95 vs. .94). But the base model scored 50\% on B1 and cannot follow the comparison instruction. \textbf{Logarithmic geometry is established during pretraining; behavioural expression requires instruction tuning.}

\textbf{E3: Bridge to SDT.} Sensitivity (\emph{d'}) from B1 forced-choice data increased with magnitude ratio for Llama (mean CV across baselines = 0.287), consistent with partial Weber constancy.

\textbf{E5: Early layers functionally implicated, late layers not causally engaged.} Repeating activation patching with B1-format (approximate) prompts (baseline accuracy 78.6\%) revealed a clear dissociation:

\emph{Table~\ref{tab:causal}. Causal intervention with approximate (B1-format) prompts. Specificity = magnitude-direction $|$ $\Delta$p$|$ / random-direction $|$ $\Delta$p$|$.}

\begin{table}[t]
\caption{Causal intervention with approximate (B1-format) prompts. Specificity = magnitude-direction $|\Delta p|$ / random-direction $|\Delta p|$.}
\label{tab:causal}
\centering
\footnotesize
\begin{tabular}{p{2.2cm}p{2.5cm}p{2.5cm}p{2.5cm}p{2.5cm}}
\toprule
& \textbf{Llama L5 (early)} & \textbf{Llama L23 (peak RSA)} & \textbf{Mistral L13 (early)} & \textbf{Mistral L22 (peak RSA)} \\
\midrule
Baseline accuracy & 0.786 & 0.786 & 0.744 & 0.744 \\
Mag mean $|$ $\Delta$p$|$ & 0.028 & 0.001 & 0.011 & 0.001 \\
Specificity ratio & 4.07$\times$ & 1.22$\times$ & 2.85$\times$ & 2.29$\times$ \\
Dose-response & Monotonic $\downarrow$ & Flat & Monotonic $\downarrow$ & Flat \\
\bottomrule
\end{tabular}
\end{table}

At early layers where RSA correlations are modest, patching along the magnitude direction produces causally specific, dose-dependent shifts. At late layers where RSA is strongest, patching is ineffective. \textbf{The magnitude subspace that is most geometrically structured is not the one the model uses for comparison.}

\begin{figure}[t]
\centering
\includegraphics[width=\textwidth]{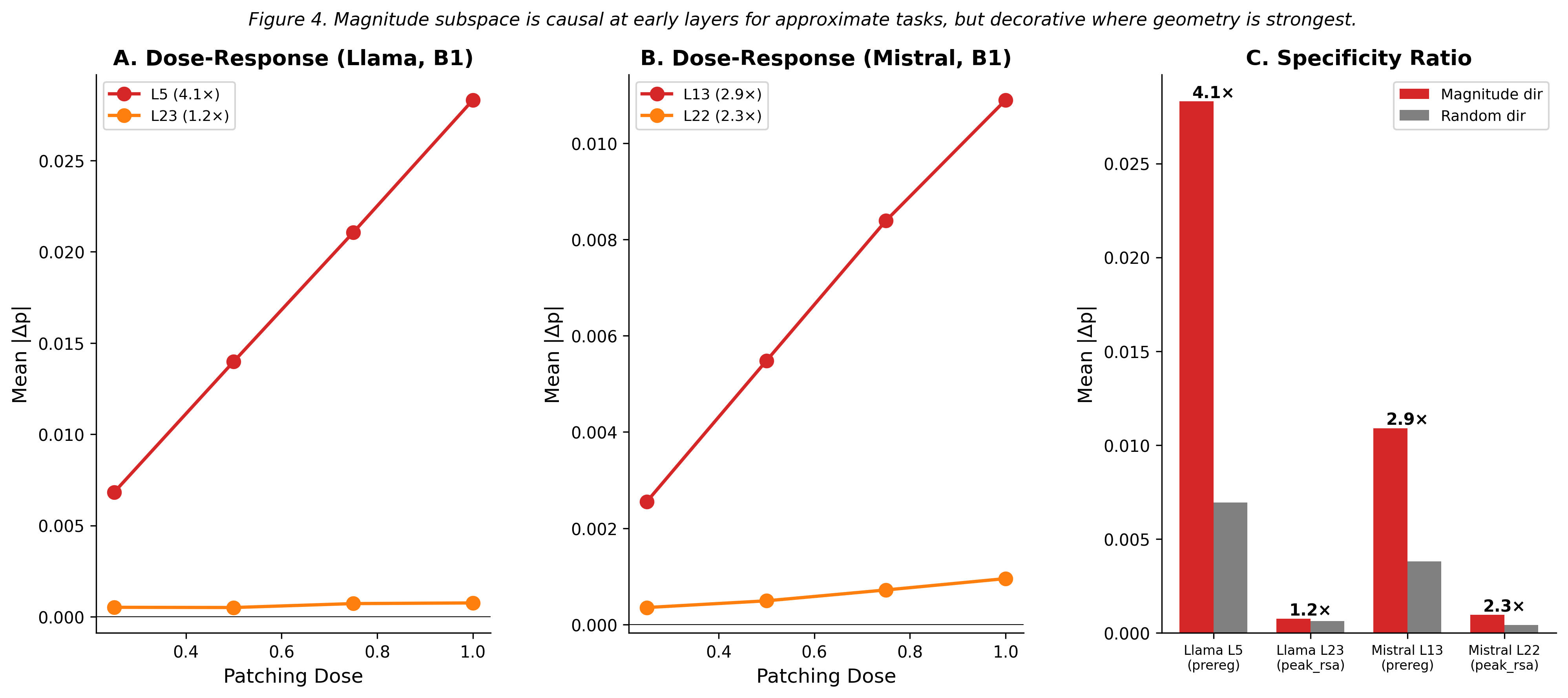}
\caption{Magnitude subspace is causal at early layers for approximate tasks, but decorative where geometry is strongest.}

\label{fig:figure4}
\end{figure}

\subsection{Robustness Controls}
\label{sec:controls}

\textbf{All six controls passed.} (1) Digit boundaries are real (Cohen's $d$ > 0.8 at most layers) but subordinate: log-magnitude partial $R^2$ = 55--63\% vs. digit-count partial $R^2$ = 8--15\%. (2) Single-token control: $\Delta$$R^2$ = .006 (tokenisation confound ruled out). (3) Unit-boundary check: equivalent-magnitude cross-unit similarity < different-magnitude same-unit similarity, confirming that geometry is form-specific and does not abstract across units. This explains the temporal/spatial behavioural null. (4) FP16--BF16 worst $r$ = .9999. (5) Frequency-matched nouns (see Supplementary Figure F7): number RDM shows structured log gradient; noun RDM is unstructured. Semantic diagnostic: noun geometry tracks meaning ($\rho$ = .41--.66 with semantic RDM), not frequency ($\rho$ = .04--.17 with frequency-rank RDM). Together, controls (1), (2), and (5) address the string-number entanglement identified by \citet{veselovsky2025what}: the logarithmic geometry is not reducible to digit structure, tokenisation, or token frequency. (6) Shuffled-magnitude: geometry tracks token identity ($\rho$ = .93), not carrier context ($\rho$ $\approx$ .00).

\begin{figure}[t]
\centering
\includegraphics[width=\textwidth]{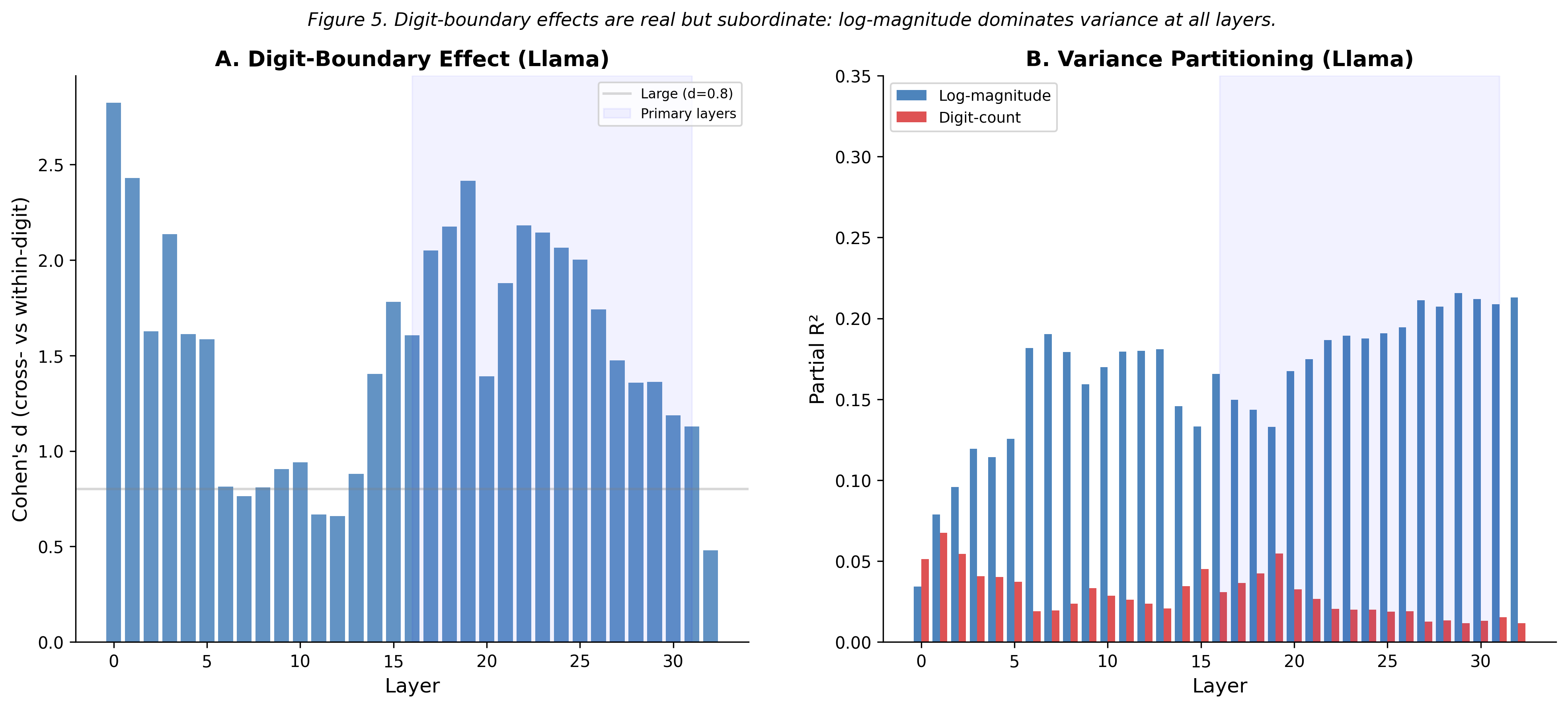}
\caption{Digit-boundary effects are real but subordinate: log-magnitude dominates variance at all layers.}

\label{fig:figure5}
\end{figure}

\begin{figure}[t]
\centering
\includegraphics[width=\textwidth]{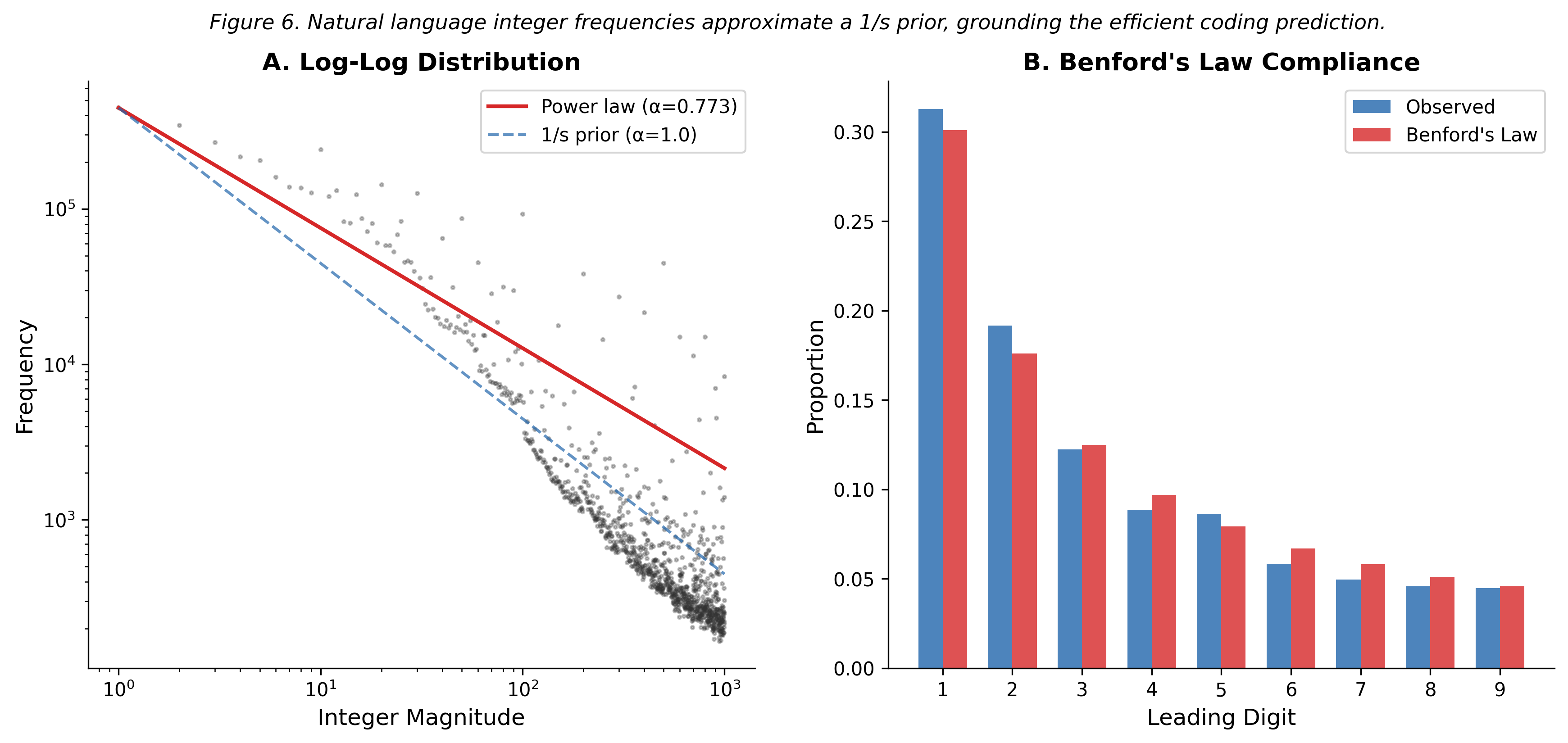}
\caption{Natural language integer frequencies approximate a $1/s$ prior, grounding the efficient coding prediction.}

\label{fig:figure6}
\end{figure}

\subsection{Exploratory Generalisation: Cross-Architecture Replication}

To test whether logarithmic geometry generalises beyond the two pre-registered model families, we ran Paradigms A, B (numerical), and C on \textbf{Qwen-2.5-7B-Instruct} (Alibaba; 7.6B parameters, 28 transformer layers, $d_{\text{model}}$ = 3584), providing a third architecture family with distinct training data and tokeniser. This analysis was conducted post-registration and is clearly labelled as exploratory. Primary hypothesis tests remain unchanged.

\textbf{Geometry.} Qwen showed logarithmic geometry at all 14 primary layers on both distance metrics (Weber $R^2$ up to .54 at mid-primary layers, RSA $\rho$ up to .79 (layer 28 showed higher values of .73 and .89, likely reflecting final-layer artefacts)). Stevens $\beta$ = 0.01, identical to both primary models. Absolute $R^2$ and $\rho$ values were somewhat lower than the primary models (which peaked at .83--.88 $R^2$ and .93--.96 $\rho$), possibly reflecting Qwen's smaller embedding dimension (3584 vs. 4096), but the qualitative pattern was identical: logarithmic preferred at every primary layer, linear never preferred.

\textbf{Behaviour.} On B1 cross-format comparison (using the identical prompts administered to the primary models), Qwen achieved 83.9\% overall accuracy, above both Llama (78.5\%) and Mistral (74.8\%). The $\Delta$ deviance test favoured log ratio over absolute difference ($\Delta$ deviance = 10.47, $p$ = .001), passing the pre-registered H2 criterion and showing a stronger effect than Llama ($\Delta$ deviance = 7.51). The aggregate Weber fraction point estimate was 0.179, within the human range (0.10--0.25), though the BCa 95\% CI was wide {[}0.037, 1.533{]} due to high overall accuracy compressing the dynamic range at most baselines. Mean logit entropy was 0.077, far below Llama's (0.288) and Mistral's (0.225), indicating highly confident, near-exact processing. Yet the ratio effect was still present, demonstrating that Weber-like ratio dependence is not contingent on approximate processing style.

\textbf{Precision gradient.} H3 was not supported (0/29 layers significant), matching Mistral's pattern and confirming that the precision gradient is not a universal property of logarithmic geometry.

\begin{figure}[t]
\centering
\includegraphics[width=\textwidth]{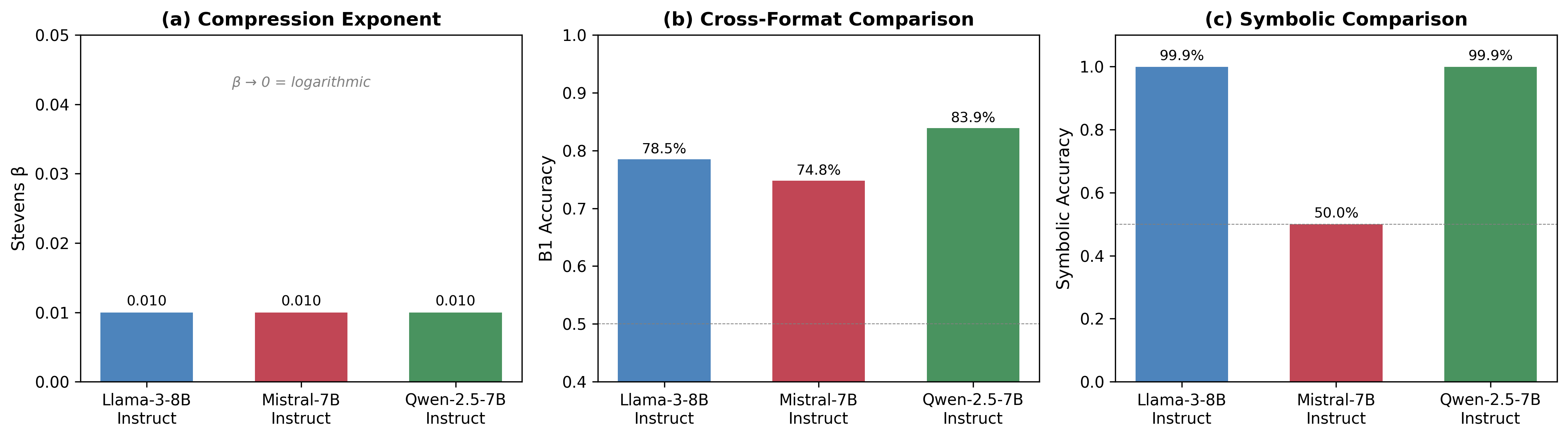}
\caption{Cross-architecture generalisation.}

\label{fig:figure7}
\end{figure}

\section{Discussion}

\subsection{Data Statistics Are Consistent with Logarithmic Geometry}

The consistency of H1 is the study's most consistent result. Across three architecturally distinct instruction-tuned models, three magnitude domains, two distance metrics, and all primary layers, representational geometry was logarithmic, with Stevens $\beta$ converging to 0.01 in every case. The addition of Qwen-2.5-7B-Instruct as an exploratory third architecture family strengthens this claim: logarithmic geometry was observed in all four models tested (three instruct, one base), spanning three architecture families (Llama, Mistral, Qwen) with distinct training data distributions. This resolves the \citeauthor{alquabeh2025number}--\citeauthor{zhu2025language} contradiction: the geometry is logarithmic at the level of representational dissimilarity (AlQuabeh), but linear probes succeed because the compression is smooth and monotonic (Zhu). Both findings are correct; they describe different aspects of the same representation.

The efficient coding interpretation is consistent with these observations. The training data's integer distribution follows a power law ($\alpha$ = 0.773), close to the scale-invariant prior that \citet{ganguli2014efficient} identify as the precondition for logarithmic compression. Gradient-based cross-entropy minimisation on this input produces the same qualitative encoding scheme, without metabolic constraints, firing-rate bounds, or population-size limits. \textbf{These results are consistent with the hypothesis that data statistics alone are sufficient, though we note the evidence is correlational: we observe a power-law input distribution and log-compressive geometry, but do not manipulate the distribution.}

But the compression observed in the models tested exceeds typical biological exponents. The Stevens exponent $\beta$ $\to$ 0.01, compared with human exponents of 0.33--3.5. Two non-exclusive explanations may account for this. First, the absence of metabolic constraints (firing-rate bounds, population-size limits) removes the floor on encoding resolution that biological systems face. Second, the shared embedding space creates a representational bottleneck: the model must compress three orders of numerical magnitude into a space that simultaneously encodes semantic, syntactic, and relational features, favouring aggressive compression. This suggests that biological Weber fractions reflect the \emph{interaction} of efficient coding with resource budgets. The same distributional statistics produce the same qualitative encoding scheme, but the quantitative compression is modulated by the system's capacity constraints. A sceptical reading of $\beta$ $\approx$ 0.01 might argue that the manifold is dominated by token-identity or lexical-frequency factors rather than magnitude per se. The variance partitioning analysis weighs against this: log-magnitude accounts for 55--63\% of unique RDM variance compared with 8--15\% for digit-count, and the frequency-matched noun control produces near-zero RSA, indicating that the log-compressive distance structure is specific to magnitude and not an artefact of token statistics. We note, however, that RSA is basis-agnostic: it identifies a log-compressive distance structure but does not uniquely determine the underlying representational format. Any smooth monotonic compression of magnitude would produce similar RSA correlations. Our claim is therefore that the distance geometry is consistently log-compressive, not that the model encodes magnitude as a scalar logarithm.

\subsection{Geometry Does Not Guarantee Competence}

The temporal/spatial behavioural null is a particularly informative dissociation. Both models possess logarithmic geometry for durations and distances (RSA correlations as strong as for numerical magnitudes) yet perform at chance on temporal/spatial discrimination.

The unit-boundary control provides a plausible mechanism: temporal and spatial stimuli use mixed units (``5 minutes'', ``2 hours''), and the model's geometry is form-specific and does not abstract across unit boundaries. ``120 seconds'' and ``2 minutes'' occupy distinct points in representational space. For numerical magnitudes, format is not an issue (``47'' is always ``47''). However, this means the temporal/spatial behavioural null may reflect a unit-conversion confound rather than a genuine absence of magnitude comparison ability: the task requires not only magnitude processing but also semantic normalisation across unit boundaries (e.g., comparing ``1 year'' with ``200 days''). The pre-registration specified a single-unit fallback (all durations in seconds, all distances in metres) to be triggered if the unit-boundary check revealed low cross-unit similarity; this fallback was not executed (see Appendix A), which limits the strength of the ``geometry without competence'' claim for temporal and spatial domains. A single-unit follow-up would disambiguate unit-conversion failure from genuine magnitude-comparison failure, though at the cost of ecological validity.

This has implications for the broader LLM interpretability literature. \textbf{Representational structure, however robust, does not imply computational use.} Probing studies that identify structured representations (linear probes, RSA, PCA) cannot, on their own, establish that the model uses those representations for downstream tasks. Our Paradigm D results reinforce this: the most geometrically structured layers are not the functionally active ones.

\subsection{The Causal Layer Dissociation}

The E5 finding deserves particular attention: early layers are functionally implicated while late layers are not causally engaged. At layer 5, where RSA correlations are modest, activation patching along the magnitude direction produces a 4.07$\times$ specificity ratio (magnitude-direction $|$ $\Delta$p$|$ = 0.028 vs. random-direction $|$ $\Delta$p$|$ = 0.007) with monotonic dose-response and correct causal sign. At layer 23, where RSA is at its peak, specificity drops to 1.22$\times$ ($|$ $\Delta$p$|$ = 0.011 vs. 0.009). The absolute effect sizes are modest (a 2.8 percentage-point shift on a 78.6\% baseline), but the specificity ratio and dose-response pattern indicate a genuine causal signal rather than noise. Two important caveats apply. First, the patching direction is derived from a degenerate probe ($R^2$ = 1.0 due to 4096 dimensions for 26 data points); although PCA validation confirmed alignment (r > .80 with PC1), the direction is not uniquely determined, and multiple independent directions were not tested. Second, patching tests sufficiency along a single linear direction, not necessity: late layers may use magnitude via nonlinear readout or distributed encoding across multiple subspaces that additive patching along one direction does not detect.

One interpretation draws on the distinction between representations that serve ongoing computation and representations that structure information for downstream prediction. Early layers may encode magnitude in an \emph{operational} format, serving as a tool used by comparison circuits to resolve the current prompt. Late layers may encode magnitude as part of a \emph{structured summary} for next-token prediction, formatting information in a way that is geometrically organised but not accessed by the comparison pathway. Under this view, the late-layer geometry is not epiphenomenal in general. It likely serves other functions such as predicting magnitude-related continuations, but it is not the substrate for the comparison computation we are testing.

This finding corroborates and extends a key observation from \citet{zhu2025language}, who noted that encoded number values on late layers are ``simply remembered but not used by the language model'' (their Section 4.1). Our causal intervention provides the mechanistic evidence for their correlational observation: late-layer representations are geometrically structured but causally inert for comparison. The convergence across independent methodologies (linear intervention in Zhu et al., activation patching in the present study) strengthens the conclusion.

This finding is methodologically important for the mechanistic interpretability community. It demonstrates that \textbf{causal relevance and representational structure can dissociate across layers,} and that the layer with the most interpretable geometry is not necessarily the layer doing the computational work. Activation patching, not probing, is required to establish functional relevance.

\subsection{Instruction Tuning as the Behavioural Switch}

The base model comparison (E1) cleanly separates the origins of geometry and competence. Base Llama-3-8B has equal or stronger magnitude geometry ($\rho$ = .95) but zero behavioural competence (50\% on B1). \textbf{Geometry is a pretraining property; behaviour is an instruction-tuning property.} This is consistent with the view that pretraining builds representational structure while fine-tuning builds task-specific readout.

\subsection{Cross-Architecture Convergence and Behavioural Divergence}

The addition of Qwen-2.5-7B-Instruct transforms the two-model divergence into a three-way pattern that is more interpretively constraining (Figure 7). All three models share identical geometry ($\beta$ = 0.01), but diverge on behaviour in a way that dissociates two dimensions: readout functionality and processing mode.

\textbf{Readout functionality.} Llama and Qwen both achieve 99.9\% on symbolic comparison, demonstrating a functional pathway from magnitude representations to comparison responses. Mistral achieves 50.0\% (position bias), demonstrating a broken readout. The magnitude representations are present in all three models (RSA geometry is equally logarithmic), so Mistral's failure is \emph{readout-specific}, not representational. We cannot fully rule out prompting artefacts or tokenisation differences (Mistral's all-multi-token number encoding may interact differently with the comparison instruction), but the same chat template and A/B labelling was used for all three models.

\textbf{Processing mode and ratio dependence.} Both models with functional readout pass the pre-registered H2 criterion: Llama ($\Delta$ deviance = 7.51, $p$ = .006, WF = 0.201) and Qwen ($\Delta$ deviance = 10.47, $p$ = .001, WF = 0.179). Both point estimates fall within the human range (0.10--0.25), though the confidence intervals are wide (Llama BCa 95\% CI {[}0.039, 0.360{]}; Qwen {[}0.037, 1.533{]}), spanning below and above the human range; the ratio effect is statistically reliable but the precision of the WF estimate is limited. Qwen's effect is stronger despite processing comparisons with near-exact confidence (entropy = 0.077 vs. Llama's 0.288). This is a noteworthy dissociation: it demonstrates that the ratio effect is not contingent on approximate, noisy processing. A model can be highly confident in its individual comparisons while still showing ratio-dependent accuracy across items. Mistral, with intermediate entropy (0.225) but a broken readout, shows no ratio effect ($\Delta$ deviance = --12.07). The critical variable for Weber-like behaviour is therefore the readout mechanism, not the processing mode.

For the efficient coding account, this three-way pattern demonstrates that logarithmic geometry is a \emph{convergent} property of transformer pretraining, appearing across all architecture families tested, but the readout mechanism that translates geometry into behaviour is determined by instruction tuning and is model-specific. The distributional structure of the input is consistent with determining the encoding; the fine-tuning procedure determines whether and how the encoding is used. We reiterate that this evidence is correlational: establishing a causal link between input distribution and representational geometry would require training on manipulated distributions, which is beyond the scope of the present study.

\subsection{Limitations}

Several limitations warrant discussion. First, although three architecture families were tested, all are 7--9B decoder-only models. Generalisability to larger models, encoder-decoder architectures, or models trained on markedly different data distributions is untested. The Qwen results partially address this concern but do not eliminate it. Second, the Paradigm D probe overfits (probe $R^2$ = 1.0 due to $p$ \textgreater> $n$: 4096 dimensions for 26 data points). Although PCA validation confirmed the direction, the probe's perfection reflects dimensionality, not encoding quality. Third, the temporal/spatial behavioural null may reflect stimulus format (mixed units) rather than a genuine absence of comparison ability; a single-unit format might yield different results, though at the cost of ecological validity. Fourth, two pre-registered robustness checks (word-form number probing and sentence-final token extraction) were not conducted and are deferred to subsequent work. Fifth, the study was massively overpowered for H1 and H2 (pre-registered Monte Carlo power $\geq$99\% at conservative effect sizes) but adequately powered for H3 ($\geq$80\% at $|$ $\rho$$|$ $\geq$ .56) and H7 ($\geq$80\% at d $\geq$ 3.0). The consistency of H1 is therefore expected given the power; the mixed results on H3--H7 are more informative. Sixth, RSA with Spearman rank correlation is basis-agnostic and cannot distinguish logarithmic encoding from other smooth monotonic compressions; the model comparison space (linear, log, power) does not include alternatives such as monotonic splines or piecewise-linear models, though these lack theoretical motivation from efficient coding. Seventh, the efficient coding interpretation is correlational: we observe a power-law input distribution and log-compressive geometry but do not manipulate the training distribution. Establishing a causal link would require counterfactual training or systematic reweighting. Eighth, \citet{pardo2019mechanistic} showed that the mechanistic foundation of Weber's Law in biological systems requires bounded exact temporal accumulation with Poisson-like variability. Transformers, operating on discrete token sequences, lack this temporal accumulation mechanism. The functional convergence at the level of encoding may reflect shared distributional statistics, while the divergence at the level of behaviour may reflect the absence of this mechanistic substrate.

\subsection{Methodological Contribution: Psychophysics as Interpretability}

Beyond the substantive findings, we propose that psychophysics offers a principled interpretability methodology for continuous representations in neural networks. The discipline provides theory-derived predictions about representational geometry (efficient coding), standardised measures of behavioural sensitivity (Weber fractions, \emph{d'}), and formal methods for testing the functional form of encoding (RSA, psychometric functions). These tools were developed over nearly two centuries to characterise the relationship between stimulus magnitudes and mental representations. This is exactly the question that arises when we ask how neural networks represent magnitude, similarity, or probability.

The four-paradigm design used here (representational geometry, behavioural discrimination, precision gradients, causal intervention) is general. It can be applied to any continuous variable that a model represents (e.g., probability for calibration, similarity for retrieval, value for reward). The psychophysical framework provides the vocabulary for describing what we find, including not just ``logarithmic'' or ``linear'', but Weber fractions, Stevens exponents, precision gradients, and the specific dissociation patterns that distinguish geometry from competence.

\section{Conclusion}

In all models tested (three instruction-tuned models spanning three architecture families, plus one base model), training on natural language produced log-compressive magnitude distance structure with Stevens $\beta$ $\approx$ 0.01, consistent with efficient coding under the distributional statistics of the training data. This geometry was robust across all models, all three domains, all primary layers, and both distance metrics. But its behavioural expression was dissociated: model-specific, domain-restricted, and (where causally tested) functionally implicated at early layers where geometry is weakest rather than at later layers where it is strongest. The critical variable determining behavioural competence is the readout mechanism established during instruction tuning, not the representational geometry established during pretraining. These findings are consistent with the hypothesis that data statistics are sufficient for the encoding; biological constraints may be necessary for the full behavioural phenotype.

\bibliography{weber}
\bibliographystyle{tmlr}

\appendix
\section{Deviations from Pre-Registration}

This study was pre-registered on OSF prior to data collection \url{https://osf.io/5r76n/overview?view_only=249cfdc140f9490da362d12fac81a7e5}. The following deviations occurred.

\textbf{Prompt formatting.} Explicit A/B labels and the model's chat template were added to Paradigm B prompts after the pre-registered format produced 50\% accuracy with position bias. Stimulus content (expressions, ratios, baselines) was unchanged. The pre-registered-format result is reported as a methodological finding: forced-choice psychophysics with instruct LLMs requires explicit option labelling.

\textbf{Cross-precision reference.} BF16 replaced FP32 (VRAM constraint + ROCm kernel errors). BF16 provides a scientifically stronger comparison (different precision--exponent tradeoff). Documented in v2.8 amendment.

\textbf{Frequency-matching.} Multi-prompt averaging (5 prompts) replaced single-prompt estimation; Hungarian algorithm replaced greedy rank-matching. Gate criterion ($\rho$ > .85) retained and met (.89). Documented in v2.8.

\textbf{Probe overfitting.} Paradigm D linear probe achieved $R^2$ = 1.0 at all layers (4096 dimensions, 26 data points). Layer selection via probe $R^2$ was degenerate. PCA validation confirmed direction alignment; E5 used multiple layers.

\textbf{Unit-boundary fallback not executed.} The pre-registration specified that if the unit-boundary manipulation check revealed low cross-unit similarity (cosine < 0.70), temporal and spatial analyses should switch to a single-unit format as the primary condition. The check confirmed form-specific geometry (equivalent-magnitude cross-unit similarity < different-magnitude same-unit similarity), triggering the fallback. However, the single-unit format was not run. Rationale: both models performed at chance (47--50\%) on temporal and spatial B1, demonstrating that these domains are behaviourally compromised regardless of format. The single-unit fallback would not have changed the programme-level H2 verdict. The geometry results (Paradigm A) are unaffected because they do not depend on the comparison format. This is deferred to subsequent work.

\textbf{Word-form number probing and sentence-final token extraction not conducted.} These were pre-registered as secondary robustness checks for Paradigm A. Word-form numbers (``forty-seven'' vs. ``47'') would test format-independence of the geometry; sentence-final token extraction would test position-dependence. Neither was run. Given the strength of H1 (all layers, all domains, both metrics, three architecture families), and the existing controls (single-token, shuffled-magnitude, frequency-matched nouns), these omissions are unlikely to alter conclusions. Deferred to subsequent work.

\section{Reproducibility Statement}

All hypotheses, analysis plans, success criteria, and stimulus specifications were pre-registered on OSF prior to data collection \url{https://osf.io/5r76n/overview?view_only=249cfdc140f9490da362d12fac81a7e5}. All analysis code, stimulus files (generated deterministically from seed 42), raw results JSONs, and figure generation scripts are archived at \url{https://github.com/synthiumjp/weber}. Model commit hashes for HuggingFace Hub are recorded in the repository. The complete experiment (stimulus generation through figure production) runs in under 30 minutes per model on a single AMD Radeon RX 7900 GRE GPU (16 GB VRAM).

\textbf{Open science commitments.} All results are reported regardless of outcome. Null results on H2, H3, H5, H6, and H7 are reported with the same detail as the positive H1 result. Deviations from the pre-registration are documented in Appendix A. The pre-registration was submitted before any data were collected or analysed.

\end{document}